\documentclass[sn-mathphys,Numbered]{sn-jnl}

\usepackage{graphicx}%
\usepackage{multirow}%
\usepackage{amsmath,amssymb,amsfonts}%
\usepackage{amsthm}%
\usepackage{mathrsfs}%
\usepackage[title]{appendix}%
\usepackage{xcolor}%
\usepackage{textcomp}%
\usepackage{manyfoot}%
\usepackage{booktabs}%
\usepackage{algorithm}%
\usepackage{algorithmic}%
\usepackage{listings}%
\usepackage{indentfirst}
\theoremstyle{thmstyleone}%
\theoremstyle{thmstyletwo}%

\theoremstyle{thmstylethree}%

\begin{document}

\title[Article Title]{A Non-Uniform Low-Light Image Enhancement Method with Multi-Scale Attention Transformer and Luminance Consistency Loss}




\author[1]{\fnm{Xiao} \sur{Fang}}\email{fang@bupt.edu.cn}

\author*[1]{\fnm{Xin} \sur{Gao}}\email{xlhhh74@bupt.edu.cn}

\author[2]{\fnm{Baofeng} \sur{Li}}\email{libaofeng@epri.sgcc.com.cn}

\author[2,3]{\fnm{Feng} \sur{Zhai}}\email{zhaifeng@epri.sgcc.com.cn}

\author[2]{\fnm{Yu} \sur{Qin}}\email{qinyuqqwwe@126.com}

\author[1]{\fnm{Zhihang} \sur{Meng}}\email{mengzhihang@bupt.edu.cn}

\author[4]{\fnm{Jiansheng} \sur{Lu}}\email{lujiansheng@sx.sgcc.com.cn}

\author[4]{\fnm{Chun} \sur{Xiao}}\email{tyutxiaochun@163.com}

\affil[1]{\orgdiv{School of Artificial Intelligence}, \orgname{Beijing University of Posts and Telecommunications}, \orgaddress{\city{Beijing}, \postcode{100876}, \country{China}}}
\affil[2]{\orgdiv{China Electric Power Research Institute Company Limited}, \orgaddress{\city{Beijing},\postcode{100192}, \country{China}}}
\affil[3]{\orgdiv{School of Electrical and Information Engineering}, \orgname{Tianjin University}, \orgaddress{\city{Tianjin}, \postcode{300072}, \country{China}}}
\affil[4]{\orgdiv{State Grid Shanxi Marketing Service Center}, \orgaddress{\city{Taiyuan}, \postcode{030032}, \country{China}}}


\abstract{Low-light image enhancement aims to improve the perception of images collected in dim environments and provide high-quality data support for image recognition tasks. When dealing with photos captured under non-uniform illumination, existing methods cannot adaptively extract the differentiated luminance information, which will easily cause over-exposure and under-exposure. From the perspective of unsupervised learning, we propose a multi-scale attention Transformer named MSATr, which sufficiently extracts local and global features for light balance to improve the visual quality. Specifically, we present a multi-scale window division scheme, which uses exponential sequences to adjust the window size of each layer. Within different-sized windows, the self-attention computation can be refined, ensuring the pixel-level feature processing capability of the model. For feature interaction across windows, a global transformer branch is constructed to provide comprehensive brightness perception and alleviate exposure problems. Furthermore, we propose a loop training strategy, using the diverse images generated by weighted mixing and a luminance consistency loss to improve the model’s generalization ability effectively. Extensive experiments on several benchmark datasets quantitatively and qualitatively prove that our MSATr is superior to state-of-the-art low-light image enhancement methods, and the enhanced images have more natural brightness and outstanding details. The code is released at https://github.com/fang001021/MSATr.}




\maketitle

\section{Introduction}\label{sec1}

Due to unavoidable environmental or technical limitations, the quality of images captured in low-light conditions is often displeasing. It can negatively impact subsequent tasks such as image classification\citep{bib1,bib2}, target detection\citep{bib3,li2021automatic,chen2021gpsd,huang2023detecting} and image generation\citep{jiang2022photohelper,sheng2021improving,xie2021bagfn,cui2023fusing}. Therefore, low-light image enhancement (LLIE) aims to improve the quality of images acquired in poor-lighting environments. It helps to enhance the visual quality, and the enhanced images can also be used for subsequent advanced vision tasks. 

In the early research on low-light image enhancement, traditional methods\citep{bib4,bib5} mainly use artificially designed models and filter structures. However, problems like color bias and insufficient details often occur when processing low-light images. It is also tricky to flexibly make adaptive adjustments in the face of diverse and complex scenarios. In recent years, deep learning-based LLIE methods have developed rapidly, relying on their excellent deep feature extraction capabilities to improve image restoration performance without difficult manual parameter adjustment. Most deep learning methods\citep{bib6,bib7,bib8}learn feature representations from pairs of images in a supervised manner, requiring images of different brightness in the same scene during the training process. However, meeting such stringent data requirements in actual application scenarios is complex, and an overly strict supervision process may also cause over-fitting problems. Therefore, many unsupervised methods have been proposed to complete low-light image enhancement tasks. Some methods\citep{bib9,bib10} use GAN\citep{bib11} structures to distinguish the difference between low-light and normal image sets, learning brightness enhancement rules and optimizing detail representation, while other unsupervised methods\citep{bib12,bib13} constrain features such as brightness and color through a large number of no-reference loss functions.

\begin{figure}[h]%
\centering
\includegraphics[width=0.8\textwidth]{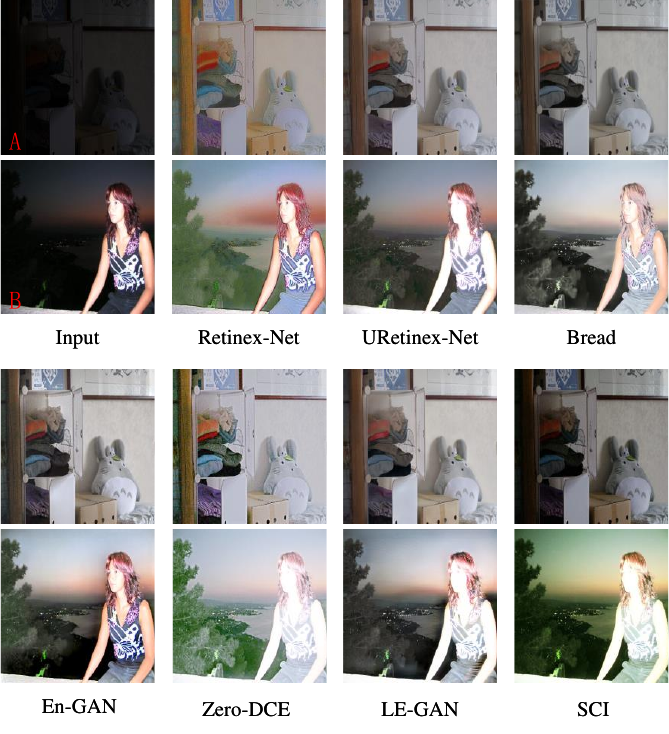}
\caption{Test images enhanced by existing methods. Input A is a relatively uniform low-illumination picture. Input B is a non-uniform picture with significant regional illumination differences.
}\label{fig1}
\end{figure}

However, whether it is a supervised or unsupervised method, how to deeply fit low-light image data features from limited training data, adaptively process diverse images in natural scenes, and balance image brightness is still a vital issue in this visual task. As shown in Figure \ref{fig1}, when processing low-light images with relatively balanced illumination in input A, most existing deep-learning models can effectively improve visual quality by enhancing brightness, improving detail representation, and reducing noise. However, in the real world, due to differences in the reflection properties of different objects or the influence of fill-light equipment such as flash, uneven low-light images similar to input B are very common, in which the character is in the highlighted area, and the background is almost entirely dark. When enhancing this type of image, many methods may have difficulty adaptively perceiving the brightness differences in different regions, resulting in over-enhancement of bright areas and under-enhancement of dark spots.

In response to the above problems, some low-light image enhancement algorithms \citep{bib10,bib14} use attention calculations during the convolutional encoding and decoding processes to alleviate over- and under-enhancement to a certain extent. However, in recent years, literature\citep{bib15,bib16} has demonstrated the inevitable loss of information during the convolution compression process, which may make the brightness characteristics and structural information more challenging to capture. In the Vision Transformer proposed in 2021, Dosovitskiy et al. \citep{bib17} guided image classification by slicing the image into blocks and combining attention features between image blocks. Compared with convolutional attention, the transformer columnar structure for images can better handle long-distance dependencies, avoid information loss, and achieve better experimental results under multiple visual tasks. However, in low-light image enhancement tasks, considering complex scenes such as non-uniform lighting, how to use the columnar structure of the image transformer to extract features from limited image information better, avoid information loss while ensuring global brightness balance, and enhance the local detail representation of the image is a great difficulty in improving the attention-based low-light image enhancement process. 

Therefore, we propose a multi-scale attention Transformer named MSATr for nun-uniform low-light image enhancement. This model performs learning through generative adversarial networks independent of training data pairs. Through the local-global features extraction network, MSATr performs multi-scale window division and self-attention calculation, enhancing pixel-level detail information inside the window and the feature interaction across different windows, to achieve the overall brightness balance and finer details performance. To further solve the image exposure problems, we propose a new loop training process and an uneven loss to help the model deeply understand the complex lighting conditions in various natural scenes, thereby achieving adaptive enhancement of non-uniform low-light images and avoiding overexposure.  

In summary, the main contributions of our paper are summarized as follows:

\begin{itemize}
    \item A multi-scale attention Transformer named MSATr is presented for low-light image enhancement to solve the over-exposure and under-exposure problems. MSATr refines local feature processing through self-attention computation within multi-scale windows, ensuring finer details in enhanced images and reducing information loss. The designed local-global transformer branch network strengthens the fusion of features across different regions. It also limits the overall brightness based on multi-level regional information, generating a more natural enhanced image. 
\end{itemize}

\begin{itemize}
    \item A consistency loop training strategy is proposed to improve the model's adaptability under limited low-light data. During the model training, pairs of non-uniform images are generated by weighted mixing across photos with different brightness levels. These mixed pictures provide new and diverse references, improving the model's illumination balancing ability and generalization performance. In addition, the designed luminance consistency loss can more effectively constrain the loop process and accelerate training convergence.
\end{itemize}

\begin{itemize}
    \item Comprehensive experiments on several benchmark datasets are conducted to consistently endorse the superiority of MSATr. The results are measured in terms of visual quality and multiple image quality assessment metrics. In contrast to existing low-light image enhancement approaches, MSATr proves to be particularly adapted to nun-uniform brightness balancing and detail enhancement.
\end{itemize}

\section{Related work}

This section reviews related research work on low-light image enhancement, which mainly includes traditional and deep learning methods, and then introduces our motivation for using the transformer structure.

\subsection{Traditional methods}

Histogram equalization (HE) \citep{bib18} is a commonly used simple and fast method in traditional low-light image enhancement tasks. Researchers have proposed a variety of enhancement algorithms based on global or local histograms (contrast-limited adaptive histogram equalization \citep{bib19}, dualistic sub-image histogram equalization method \citep{bib20}, adaptive histogram equalization\citep{bib21} and brightness bi-histogram equalization method\citep{bib22}) to adjust the pixel value range of the input image dynamically. Lee et al. \citep{bib4} proposed LDR, which uses the connection between adjacent pixels and the global gray level difference to adjust the brightness of local areas, thereby achieving better visual effects locally and globally. However, these traditional HE methods cannot analyze the characteristics of the image itself, and adjustments based on the pixel value range may lead to the loss of crucial information in the picture, resulting in distortion. Therefore, some traditional algorithms decompose low-quality images' illumination and reflection components based on Retinex theory \citep{bib23} and perform brightness estimation and restoration. Jobson et al. \citep{bib24} first proposed the single-scale Retinex algorithm (SSR), using Gaussian blur input as the illumination map to solve the reflection component, and then optimized the multi-scale Retinex algorithm \citep{bib25} (MSR) on this basis, achieving better results by fusing multiple Gaussian functions with different variances. In addition to the methods of solving the reflection component, Xiao et al. \citep{bib5} proposed the LIME model, which directly finds the maximum intensity of each pixel in three channels to construct and optimize the illumination map, improving image quality and computational efficiency. However, due to the relatively fixed algorithm framework, these traditional methods often fail to achieve the expected enhancement effects in some aspects. They cannot consider the detailed information and color information of the image itself in the process of enhancing brightness. At the same time, manual adjustment of parameters is required when facing complex low-light scenes and diverse low-light images, which results in fewer practical applications in multiple scenarios.  

\subsection{Deep learning methods}

In recent years, deep learning has gradually become the leading solution for various image enhancement tasks \citep{bib26,bib27,bib28,bib29} due to its better accuracy and robustness. In low-light image enhancement, existing deep learning methods can be divided into supervised and unsupervised approaches based on data requirements. 

\subsubsection{Fully supervised methods}

Most existing low-light image enhancement methods rely on paired image data to learn differences such as brightness between images. The first low-light image enhancement algorithm based on deep learning, LLNet \citep{bib30}, simultaneously achieves end-to-end brightness enhancement and denoising through an improved stacked denoising encoder. After that, Lv et al. \citep{bib31} proposed an end-to-end multi-branch enhancement network MBLLEN, which extracts effective feature representations through the feature extraction module, enhancement module, and fusion module to improve model performance, completing several subtasks such as low-light image brightness enhancement and noise removal. Compared with end-to-end networks, deep learning methods based on physically interpretable Retinex theory have better enhancement performance in most cases. Wei et al. \citep{bib6} decomposed the image into reflection components and smooth illumination through a deep learning network for the first time and reconstructed the image through the enhancement network. On this basis, Wu et al. \citep{bib7} further optimized the image decomposition process and proposed a deep expansion network URetinex-Net, which realizes image enhancement through three modules and uses powerful deep model learning capabilities to simulate data-related priors. Recently, Guo et al. \citep{bib8} separated the three tasks of brightness enhancement, noise removal, and color correction, proposed the Bread model, and tried to solve the coupling relationship between noise elimination and color distortion in the brightness-chromaticity space. However, stringent paired images are required in supervised methods that rely on professional equipment and rigorous acquisition processes. In addition, the fully supervised training strategy can easily lead to overfitting of the model, and the model's generalization performance is not good enough. When faced with non-uniform low-light images in unseen multiple scenes, the supervised model may have difficulty adaptively balancing image brightness, resulting in over-exposure and under-exposure. 

\subsubsection{Unsupervised methods}

Unlike supervised methods that require paired data, unsupervised methods have received more and more attention and research in recent years due to their lower data requirements and better generalization capabilities. Jiang et al. \citep{bib9} proposed Enlighten-GAN, which used a generative confrontation method to train an unsupervised model for the first time and ensured image enhancement quality through a global-local discriminator. However, the simple generator structure cannot effectively guarantee fine image information processing. Therefore, Fu et al. \citep{bib10} built LE-GAN, trained the model in a generative adversarial manner, and combined attention calculations during the encoding and decoding process to optimize the network's feature extraction capabilities to solve noise and color deviation problems and improve visual quality. In addition to generative adversarial methods, some unsupervised methods use finely combined reference-free loss functions to constrain the image enhancement process. Since there is no need to train a discriminator, these methods reduce the number of parameters and computational overhead to a great extent and do not require additional data annotation at all. Guo et al. \citep{bib12} proposed a zero-reference deep learning method, Zero-DCE, which transforms the image enhancement task into a specific curve estimation. These curves are then used to dynamically adjust pixel values of low-light images to speed up model inference. Although discrete pixel value curve correction can significantly reduce the number of parameters, it ignores local area correlation. After that, Ma et al. \citep{bib13} proposed the self-calibration model SCI, which completes self-supervised learning of image enhancement through a multi-level illumination self-calibration module that shares weights without requiring any data labels. Compared with supervised learning methods, most unsupervised methods may have difficulty fitting the distribution characteristics of training data and processing multi-scale feature information, leading to worse detail performance and higher noise levels. Although the single discriminator or no-reference loss function of the method can better constrain the overall brightness level and various color texture information of the enhanced image, it is difficult to adaptively divide the light and dark areas and adjust the enhancement intensity for unsupervised methods when facing non-uniform low-light images.

\subsection{Vision Transformer application}

Vaswani et al. \citep{bib32} initially proposed that the Transformer structure be applied to natural language processing \citep{bib33} (NLP) and fully mine the feature correlation between data through Attention calculation and feed-forward neural network. Due to this structure's excellent global vision and attention mechanism, more and more image processing works have used the Transformer structure as the backbone network and achieved good results. Dosovitskiy et al. \citep{bib17} first proposed a Vision Transformer (ViT), which used the Transformer structure for non-overlapping image blocks and achieved higher accuracy and faster computing speed than the convolutional structure in image classification tasks. Based on the multi-scale and high-resolution characteristics of the image itself, Liu et al. \citep{bib34} proposed the Swin-transformer structure, which reduces model parameters and calculation volume by dividing windows and window shifts. In recent years, researchers\citep{lin2021eapt,li2023trajectory} have applied the vision transformer structure to image reconstruction tasks, verifying the feasibility of the structure and better feature processing capability than convolutional networks. Liang et al. \citep{bib35} proposed SwinIR, using serial Swin modules to perform multiple image restoration tasks, including image denoising, image compression, etc. Deng et al. \citep{bib16} improved the positional encoding process of ViT. They built an encoder and decoder based on the Transformer structure, which surpassed the existing convolutional structure-based methods in style transfer. 

To better extract features and reduce information loss, this paper introduces the vision transformer structure for low-light image enhancement. Its attention mechanism can also effectively balance brightness and reduce color deviation and noise interference. However, unlike other image tasks, low-light image enhancement requires pixel-level feature extraction capabilities to ensure the detailed representation in enhanced images. This may need a more refined transformer network. At the same time, the significant feature differences in light and dark areas of low-light shots also challenge the attention calculation process. Therefore, how to improve detailed feature-extraction capabilities and better balance differentiated light through attention computation is a great challenge for low-light image enhancement under the transformer framework.

\section{Method}

This section first introduces the MSATr's overall network structure and training process, then shows the critical local-global attention network, and finally lists several essential loss functions used in the training process.

\subsection{Network architectures}

\begin{figure}[h]%
\centering
\includegraphics[width=1.0\textwidth]{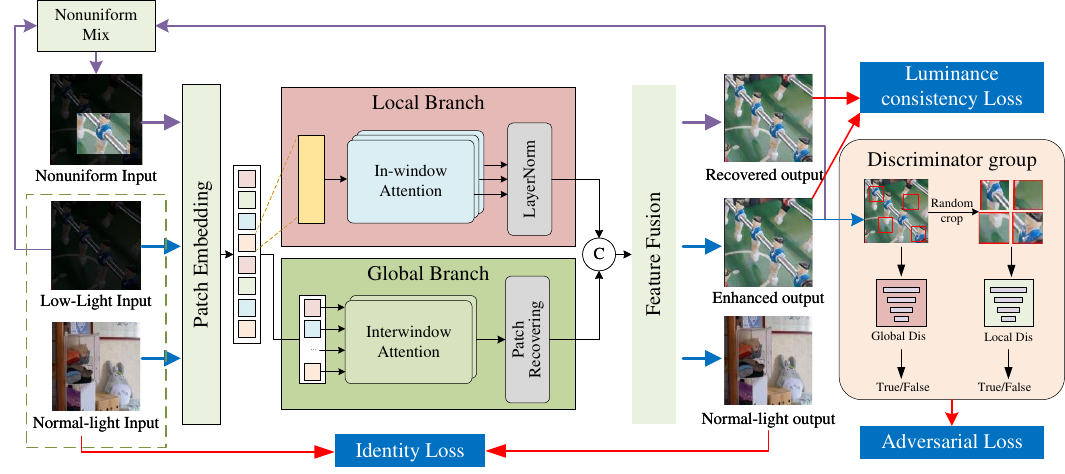}
\caption{The overall model framework. The training mainly relies on the generative adversarial and the consistency cycle process.
}\label{fig2}
\end{figure}

As shown in Figure \ref{fig2}, in the generative adversarial network, MSATr is the generator to complete the end-to-end low-light image enhancement task. The model inputs a three-channel RGB low-illumination image, maps the data to a high-dimensional feature space through a local-global network, and calculates the image's multi-feature representation based on multi-head attention\citep{bib32}. Then, multiple convolutions are used to unify the feature dimensions and merge channel features. Finally, the obtained local-global elements are fused through a convolutional network to regenerate the enhanced image. 

To better train the local-global network, the adversarial process refers to the dual discriminator structure of Enlighten-GAN \citep{bib9}. Among them, the global discriminator is used to determine the category of the entire image. In contrast, the local discriminator is used to identify the small patches randomly cut from the input image, which helps to enhance the brightness and detail performance of local areas. In addition, a loop training network and a luminance consistency loss function are also added to the training process. 

\subsection{Local-global attention network}

\begin{figure}[h]%
\centering
\includegraphics[width=1.0\textwidth]{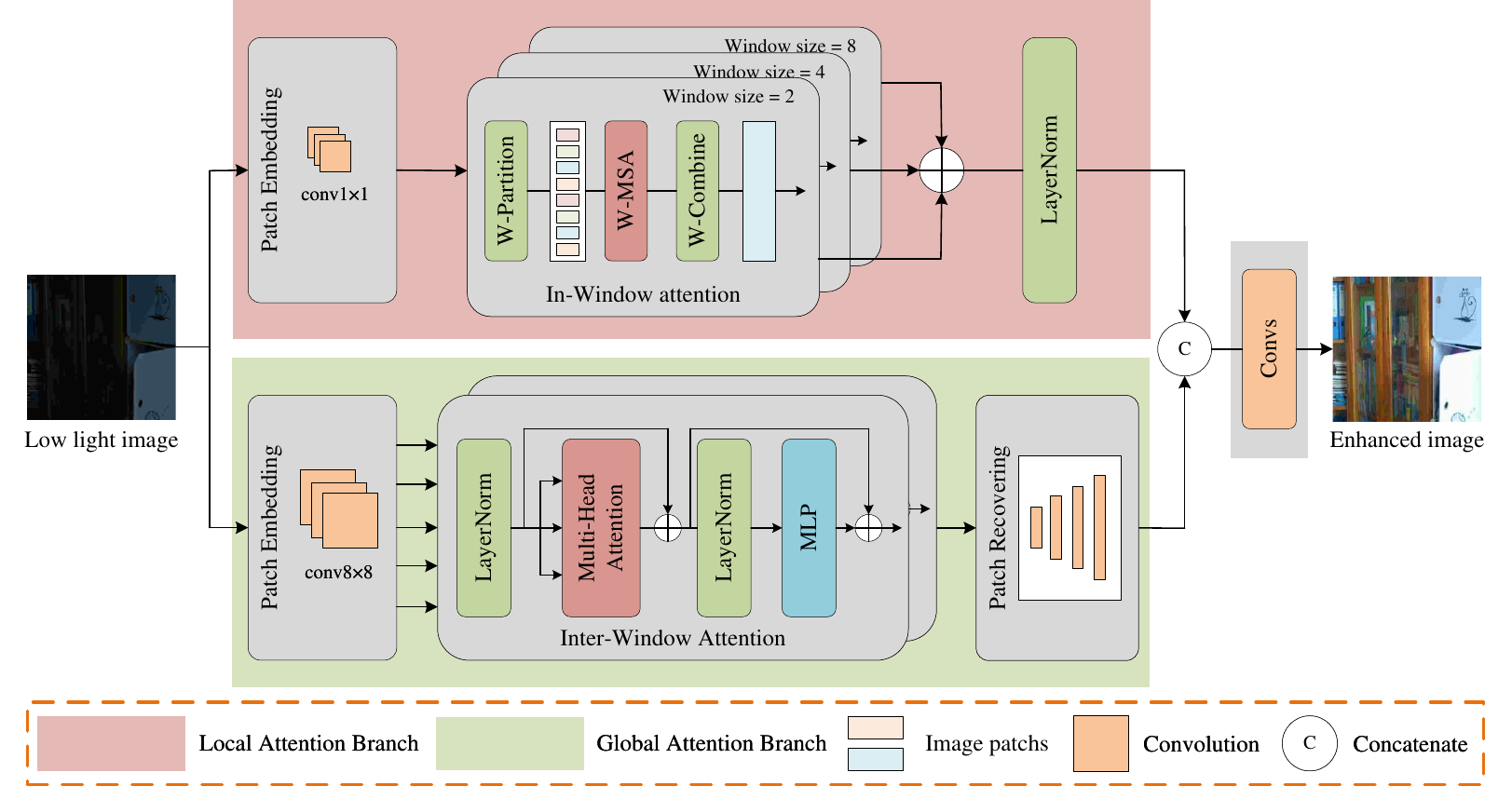}
\caption{The framework of MSATr.The network fully fuses features through local and global branches to recover low-light images.
}\label{fig3}
\end{figure}

Figure \ref{fig3} shows the low-light image enhancement network structure. The network is divided into local-global feature extraction, feature fusion, and multi-layer convolution image generation. The local-global attention network establishes pixel-level feature correlation, thereby achieving light and dark area discrimination and adaptive enhancement while maintaining good detail performance and noise levels. 

\subsubsection{Local attention feature extraction network}

The local network focuses on detailed features via in-window attention computation. To deal with brightness and color features that vary significantly between different areas, we propose a multi-scale window division mechanism, which can effectively improve the model's ability to extract diverse information. Starting with the smallest window size 2, we continuously increase the size of each i-th layer based on an exponential sequence:

\begin{equation}
Siz{{e}_{i}}={{2}^{i}}.\label{eq1}
\end{equation}

Taking into account both accuracy and efficiency, we use three multi-scale window attention layers. In different layers, a larger window can perceive more brightness information, especially for continuous changes in light and dark areas under non-uniform illumination. In contrast, for fragmented light and dark changes, a smaller window can help the model locate the light and dark dividing line and achieve better details.

\begin{algorithm}[!htb]
    \caption{Multi-scale local window self-attention computation}
    \label{a1}
    \begin{algorithmic}[1]
    \REQUIRE 
        Training datasets $T$; Number of layers ${{L}_{n}}$
    \ENSURE 
        Multi-scale mixed local feature ${{F}_{local\_m}} $
    \FOR{\textbf{each} ${{I}_{i}}$ in $T$}
	\STATE ${{I}_{AT}}\Leftarrow Embedding({{I}_{i}},1)$
  
        /*Map images into learnable feature codes and reshape*/
        \FOR{\textbf{each} $l$ in $[1,{{L}_{n}}]$}
                \IF{$l=1$}
                        \STATE ${{X}_{l}}\Leftarrow Window\ partition({{I}_{AT}},{{2}^{l}})$
                \ELSE
                        \STATE ${{X}_{l}}\Leftarrow Window\ partition({{F}_{l-1}},{{2}^{l}})$
                \ENDIF        
                \STATE ${{X}_{l}}\Leftarrow Window\ Attention({{X}_{l}})$
                \STATE ${{F}_{l}}\Leftarrow Window\ reverse({{X}_{l}})$
                
                /*Calculate the window self-attention of size l and reshape*/
                \STATE ${{F}_{local\_m,i}}\underleftarrow{+} {{F}_{l}}$
                
        \ENDFOR
    \ENDFOR
    \RETURN ${{F}_{local\_m}}$
    \end{algorithmic}
\end{algorithm}

As shown in Algorithm 1, the local attention branch first performs patch segmentation through a 1×1 convolution layer to extract pixel-level detailed features. Then, it passes through the window attention calculation module with windows 2, 4, and 8 in sequence. The window attention calculation method of W-MSA \citep{bib34} is used in each calculation module. Unlike the Swin transformer, because there is another branch for global attention calculation, MSATr abandons the original Swin window shift step to reduce computational complexity.

\subsubsection{Global attention feature processing network}

Self-attention calculation and pixel-level feature extraction within a maximum window of 8×8 have been implemented in the local branch, so the global feature processing network focuses on establishing feature connections between these windows. Through the integration of local and global information, pixel-window-pixel feature correlation can be found to achieve more detailed and comprehensive image generation. 

We first divide the image into multiple patches, each containing representative information in a large window. Operationally, the global network first performs patch segmentation through a convolutional layer with a convolution kernel size of 8×8 and a stride of 8, and then projects input patches into a sequential feature embedding $\varepsilon $. Given the input embedding sequence ${{Z}_{l}}\text{ }=\left\{ {{\varepsilon }_{l1}}\text{ }+{{P}_{l1}},{{\varepsilon }_{l2}}\text{ }+{{P}_{l2}},...,{{\varepsilon }_{lL}}+{{P}_{lL}} \right\}$of length $L$ and positional encoding ${{P}_{l}}$, two multi-head self-attention modules in series are used to establish the global connection across windows. The input sequence is encoded into query ($Q$), key ($K$), and value ($V$): 

\begin{equation}
\text{Q = }{{\text{Z}}_{l}}{{\text{W}}_{q}}\text{, K = }{{\text{Z}}_{l}}{{\text{W}}_{k}}\text{, V = }{{\text{Z}}_{l}}{{\text{W}}_{\text{v}}},\label{eq2}
\end{equation}
where ${{\text{W}}_{q}}$,${{\text{W}}_{k}}$,${{\text{W}}_{\text{v}}}$ are parameter matrices. The multi-head attention is then calculated by

\begin{equation}
{{\mathbb{F}}_{\text{MSA}}}\text{(Q,K,V) = Concat(}{{\text{Attention}}_{\text{1}}}\text{(Q,K,V), }\text{. }\text{. }\text{. , }{{\text{Attention}}_{\text{N}}}\text{ (Q,K,V))},\label{eq3}
\end{equation}
where $N$ is the number of attention heads.

To unify local and international features for subsequent feature fusion, the global feature vector dimensions are adjusted through Patch\_Recovering, which uses multi-layer upsampling and deconvolution. Ultimately, the network dimensionally stacks the extracted local attention features with the global attention features, performs feature fusion through the convolutional network, and generates a three-channel RGB image.

\subsection{Consistency loop training process}\label{3.3}

\begin{figure}[h]%
\centering
\includegraphics[width=0.8\textwidth]{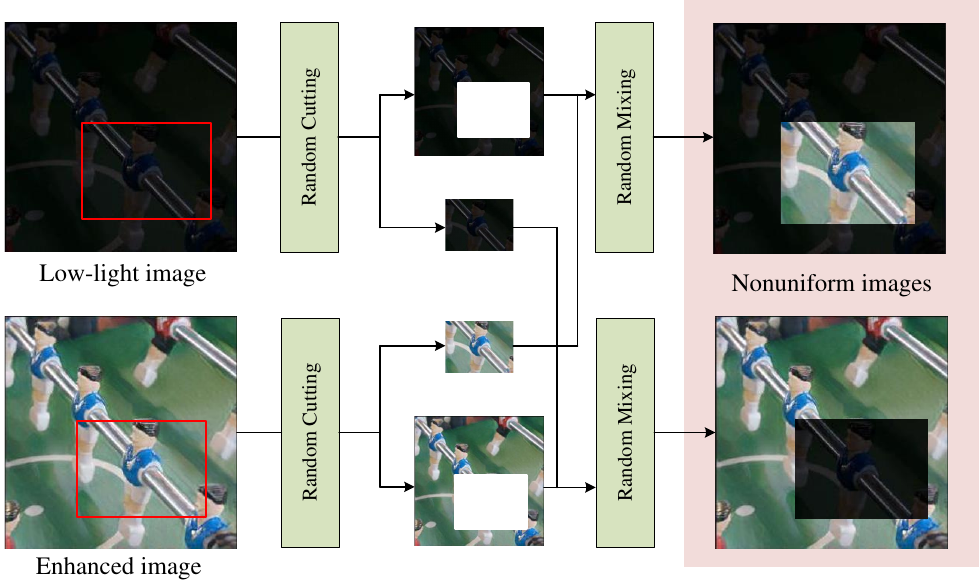}
\caption{non-uniform mixing process. the non-uniform images are generated by the random region mixing of the input image and the enhanced image for model training
}\label{fig4}
\end{figure}

To solve the problem of over-enhancement and under-enhancement, we generate diverse non-uniform data by random splicing and mixing across images. Relying on the newly developed data, the loop training process improves the model's adaptive ability and generalization performance. As shown in Figure \ref{fig4}, new training data can be obtained in each round of cycle training by mixing the input low-light image with the enhanced image of the model. In the process of image synthesis, a random region is first cropped in the low-light image and the enhanced image. Then, a random weighted mixture is carried out in the delimited region according to the following formula:
\begin{equation}
I'=\alpha {{I}_{out}}+(1-\alpha ){{I}_{in}}\label{eq4},
\end{equation}
where, ${{I}_{out}}$ and ${{I}_{in}}$ represent the enhanced image and the input image respectively. $\alpha$ is a random magnification with a size of 0~1. The larger the value, the closer the random area is to the enhanced image.

After this, the obtained uneven image data is re-input into the model, and the image can be obtained after secondary enhancement. In the best case, the obtained secondary enhanced image should be consistent with the first enhanced image to improve the model's ability to adjust the brightness. At the same time, more data modes improve the model's generalization performance.

\subsection{Loss functions}

This section mainly introduces the loss functions involved in the training process. First, a luminance consistency loss used during circuit training is proposed. This section also introduces adversarial loss, identity-invariant loss, and Self-feature-preserving loss.

\subsubsection{Luminance consistency loss}

According to Section \ref{3.3}, we synthesize new image data for loop training to improve the model's ability to process uneven images. To constrain the model training process and strengthen the model's perceptive ability, the luminance consistency loss function is proposed as follows:
\begin{equation}
{{\mathcal{L}}_{U}}=\frac{1}{\alpha mn}\mathop{\sum }_{i=1}^{m-1}\mathop{\sum }_{j=0}^{n-1}{{[I\left( i,j \right)-K\left( i,j \right)]}^{2}},\label{eq5}
\end{equation}
where $I$ and $K$ represent the restored image and the enhanced image. $mn$ indicates the area of a randomly cut image patch.

The above loss function can reasonably constrain the consistency of the image after two enhancements, thus strengthening the light balance ability of the model. In addition, the pixel-based vital supervised learning process can also accelerate the convergence of the model.

\subsubsection{Adversarial loss}

To impose constraints on the MSATr generation network and complete the task of low-light image enhancement, the adversarial process uses a local-global dual discriminator for unsupervised learning of the network. The performance of the enhanced network is evaluated through the judgment of the discriminator network on the generated images, encouraging the generated images to be closer to the distribution of normal real images. 
The global discriminator and generator losses are:
\begin{equation}
\mathcal{L}_{D}^{Global}=\log ({{D}^{Global}}({{x}_{r}}))+\log (1-{{D}^{Global}}({{x}_{f}})),\label{eq6}
\end{equation}
\begin{equation}
\mathcal{L}_{G}^{Global}=-\log (1-{{D}^{Global}}({{x}_{f}})),\label{eq7}
\end{equation}
where ${D}^{Global}$ represents the global discriminator. ${x}_{r}$ and ${x}_{f}$ represent the sampled real normal pictures and enhanced pictures.

The local discriminator and generator losses are similar to the above:
\begin{equation}
\mathcal{L}_{D}^{Local}=\log ({{D}^{Local}}(x_{r}^{patch}))+\log (1-{{D}^{Local}}(x_{f}^{patch})),\label{eq8}
\end{equation}
\begin{equation}
\mathcal{L}_{D}^{Local}=-\log (1-{{D}^{Local}}(x_{f}^{patch})),\label{eq9}
\end{equation}
where $x_{r}^{patch}$ and $x_{f}^{patch}$  represent the local area generated by random cutting from the picture. 

\subsubsection{Self-feature preserving loss}

To constrain the enhanced image to be consistent with the content information of the input low-light image, the perceptual loss proposed by Johnson et al. \citep{bib36}was referred to during the model training process. We use the pre-trained VGG model to extract feature maps to compare the content similarity of two pictures. The self-feature preserving loss can be expressed as:
\begin{equation}
{{\mathcal{L}}_{C}}=\frac{1}{{{N}_{l}}}\sum\limits_{i=0}^{{{N}_{l}}}{||{{\phi }_{i}}(G({{x}_{l}}))-{{\phi }_{i}}({{x}_{l}})|{{|}_{2}}},\label{eq10}
\end{equation}
where, ${{\phi }_{i}}(\cdot )$represents the features extracted from the i-th layer of the vgg16 network, ${{N}_{l}}$represents the number of layers, ${{x}_{l}}$represents the input low-light image and $G({{x}_{l}})$ represents the image generated by the enhancement network. 

\subsubsection{Identity invariant loss}

To avoid over-enhancement and speed up the model's fitting and learning convergence of normal images, normal images are input to the enhancement network during the model training process, and the following identity-invariant loss is used to constrain the enhancement output:
\begin{equation}
{{\mathcal{L}}_{C}}=||G({{x}_{r}})-{{x}_{r}}|{{|}_{2}},\label{eq11}
\end{equation}
where, ${x}_{r}$ represents the use of real normal images as network input, $G({{x}_{r}})$ represents the image generated by the enhancement network, and the identity-invariant loss encourages the enhancement network to maintain the brightness level of normal images, thereby avoiding over-exposure and information loss. 

\section{Experiments}

This section defines the MSATr's implementation details and parameter settings and then introduces the supervised and unsupervised data sets and indicators used. This section also conducts extensive comparative experiments between MSATr and existing advanced deep learning methods and verifies the model's excellent low-light enhancement performance and generalization ability on multiple data sets. Finally, we further demonstrate each module's impact on model performance through ablation experiments.

\subsection{Implementation details}

The model is implemented using Pytorch and optimized using the Adam optimizer, where,  ${{\beta }_{1}}$ and ${{\beta }_{2}}$ are set to 0.9 and 0.999. The initial value of the learning rate is 5e-5 and continues to decay with the training process. During the training process, in each round, batch-size low-light and normal-light pictures are randomly selected from the data set, and a 256*256 part is randomly cropped from each picture for training. The test image will be adjusted to 512*512 as input during the test. All training and testing were performed on an NVIDIA 3090ti GPU.

\subsection{Datasets and metrics}

To verify the effectiveness of the method proposed in this article, this section uses some public data sets to evaluate the model's performance. Among them, the LOL data set \citep{bib6} contains 485 pairs of low-light and normal-light training images and 15 pairs of test images. In addition, we also used five reference-free natural dark image data sets, DICM \citep{bib1}, \citep{bib5}, \citep{bib37}, \citep{bib38}, and VV\footnote{ https://sites.google.com/site/vonikakis/datasets}, to test the generalization performance of the trained model. 

Image evaluation indicators mainly include PSNR, SSIM, LPIPS\citep{bib39}, and NIQE\citep{bib40} for evaluating enhanced image quality and model performance. The first three reference indicators require normal images for comparison and calculation. PSNR reflects the difference in pixel values between the image to be evaluated and the reference image, SSIM demonstrates the similarity in brightness, contrast, and structure between the image to be considered and the reference image, and LPIPS reflects the difference between the feature maps obtained by the feature extractor between the image to be evaluated and the reference image. NIQE is a non-reference evaluation index that reflects the difference between the image to be assessed and the artificial natural image set. The lower the value, the closer the image to be evaluated is to the realistic image. 

\begin{table}[h]
\caption{Various comparison methods and detailed information mentioned in the article}\label{tab1}%
\begin{tabular}{@{}llll@{}}
\toprule
Method	& Class	& Year	& Periodical/Conference\\
\midrule

LDR	& \multirow{2}{*}{traditional} &	2013 &	IEEE T IMAGE PROCESS \\
LIME &	&	2016 &	IEEE T IMAGE PROCESS \\ \cmidrule{1-4} 
Retinex-Net &	\multirow{3}{*}{supervised} &	2018 &	BMVC \\
Uretinex-Net & &		2022 &	CVPR \\
Bread & &		2023 &	International Journal of Computer Vision \\ \cmidrule{1-4} 
Zero-DCE &	\multirow{4}{*}{unsupervised} &	2020 &	IEEE T PATTERN ANAL \\
EnlightenGAN & &		2021 &	IEEE T IMAGE PROCESS \\
SCI		& & 2022 &	CVPR \\
LE-GAN & &		2022 &	Knowledge-Based Systems \\

\botrule
\end{tabular}
\end{table}

\subsection{Comparison with state-of-the-art methods}

We compared MSATr with several traditional methods, i.e., LDR\citep{bib1} and LIME\citep{bib5}, and several state-of-the-art deep learning methods, i.e., Retinex-Net\citep{bib6}, Uretinex-Net\citep{bib7} and Bread\citep{bib8}, Zero-DCE\citep{bib12}, EnlightenGAN\citep{bib9}, SCI\citep{bib13} and LE-GAN\citep{bib10}. The details of these methods are shown in Table \ref{tab1}. All models were fully retrained and tested using official codes in the same experimental environment.

\subsubsection{Comparative experimental results on the LOL test set }

\begin{figure}[h]%
\centering
\includegraphics[width=0.9\textwidth]{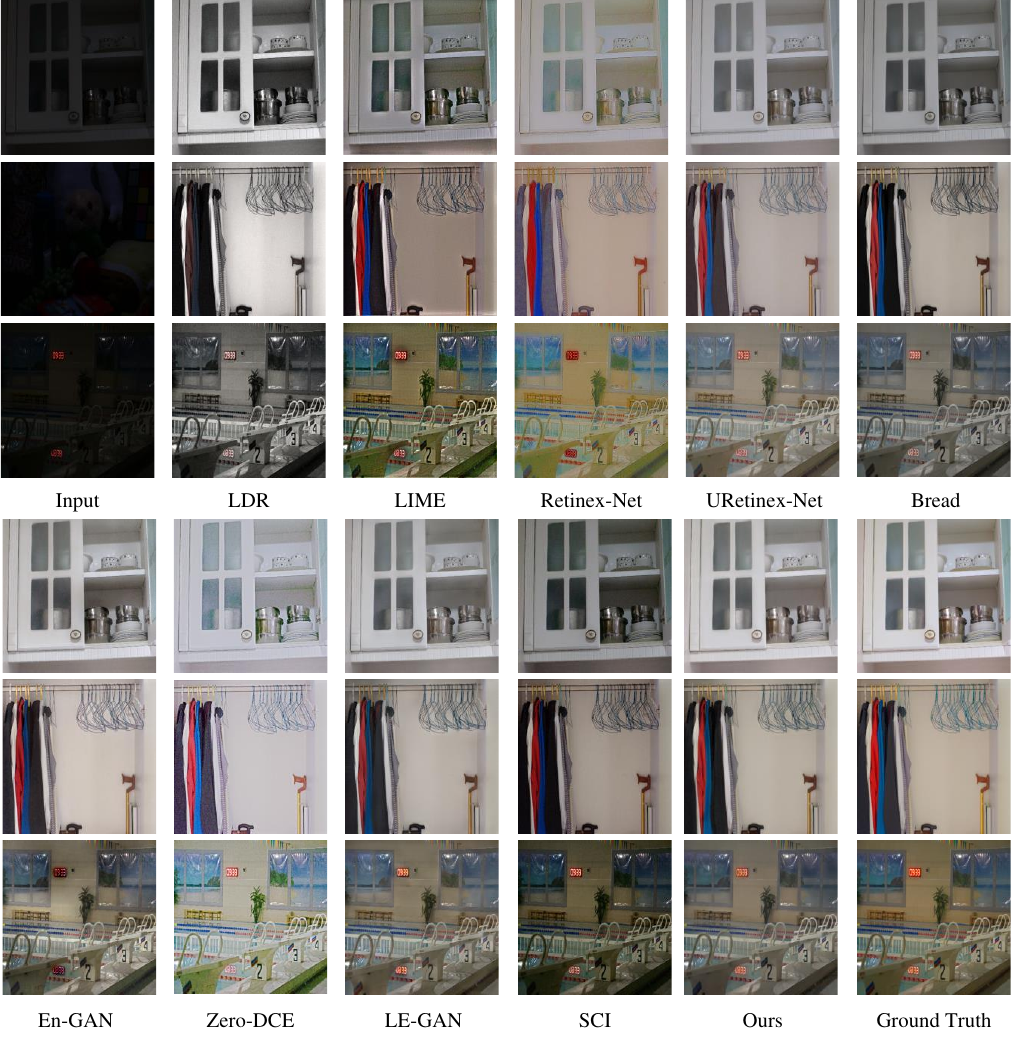}
\caption{Enhanced image comparison of each method on the LOL test set. Input shows the low-light image and Ground Truth is the normal image for comparison.
}\label{fig5}
\end{figure}

Table \ref{tab2} shows the quantitative comparison between MSATr and other competitors on the LOL test set. Our method outperforms the other unsupervised methods in all reference indicators significantly and is close to the best-supervised methods. Under the no-reference index NiQE, MSATr achieved the best score, meaning the enhanced image has a more natural and realistic visual effect. Figure \ref{fig5} shows some test image results of LOL. The image generated by MSATr has more uniform illumination levels and fewer artifacts and can consider the overall brightness, color, and local details. Most other contrast methods have problems such as insufficient enhancement, color distortion, and residual noise. 

\begin{table}[h]
\caption{PSNR, SSIM, LPIPS, and NIQE indicators of each method on LOL. The best result is in bold, the second is underlined, and the third is in italics.}\label{tab2}%
\begin{tabular}{@{}lllll@{}}
\toprule
Method	& PSNR$\uparrow$	& SSIM$\uparrow$	& LPIPS$\uparrow$ & NIQE$\downarrow$ \\
\midrule

LDR	&	15.7350	& 	0.5942	& 	0.4450	& 	4.5982	 \\
LIME	&		16.3432	& 	0.5362	& 	0.3815	& 	4.6724  \\
Retinex-Net	&	19.5082	& 	0.6779	& 	0.3735	& 	5.6239  \\
Uretinex-Net	&		19.9829	& 	\textbf{0.8398}	& 	\textbf{0.1314}	& 	4.5049  \\
Bread	&		\emph{20.0282}	& 	0.8140	& 	0.2052	& 	5.0209  \\
EnlightenGAN	&	18.8377	& 	0.7476	& 	0.2530	& 	\underline{3.5794}  \\
ZeroDCE	&		18.6882	& 	0.5649	& 	0.3999	& 	4.6757  \\
LE-GAN	&		\underline{20.9791}	& 	\emph{0.7958}	& 	\emph{0.1778}	& 	\emph{3.9806}  \\
SCI	&		14.5325	& 	0.5681	& 	0.3242	& 	5.0007  \\
MSATr	&		\textbf{21.9281}	& 	\underline{0.8141}	& 	\underline{0.1658}	& 	\textbf{3.4813}  \\

\botrule
\end{tabular}
\end{table}

\subsubsection{Generalization ability comparison}

\begin{table}[h]
\caption{NIQE indicators of each method on no reference data set. The best result is in bold, the second is underlined, and the third is in italics.}\label{tab3}%
\begin{tabular}{@{}lcccccc@{}}
\toprule
& \multicolumn{6}{c}{NIQE$\downarrow$} \\ \cmidrule{2-7}
Method &	DICM &	LIME &	MEF &	NPE &	VV &	average\\
\midrule

LDR	&	4.0293	& 	4.1941	& 	4.3400	& 	4.5533	& 	3.3588	& 	4.0951   \\
LIME	&	\underline{3.9742}	& 	\textbf{3.7396}	& 	4.1317	& 	4.5213	& 	3.7537	& 	\emph{4.0241}   \\
Retinex-Net	&	4.8139	& 	4.8568	& 	5.0660	& 	4.9521	& 	3.3391	& 	4.6056   \\
Uretinex-Net	&	4.1911	& 	4.3782	& 	4.3430	& 	4.4495	& 	\underline{3.0512}	& 	4.0826   \\
Bread	&	4.6701	& 	4.6883	& 	4.9513	& 	4.8872	& 	3.7543	& 	4.5902  \\ 
EnlightenGAN	&	\emph{3.9851}	& 	\emph{3.8719}	& 	\emph{4.0888}	& 	\underline{4.4313}	& 	\emph{3.1154}	& 	\underline{3.8985}   \\
ZeroDCE	&	4.2860	& 	3.9812	& 	4.2084	& 	4.5783	& 	3.1408	& 	4.0389  \\ 
LE-GAN	&	4.4561	& 	\underline{3.8605}	& 	\underline{3.9180}	& 	4.8127	& 	3.4531	& 	4.1001   \\
SCI	&	4.3732	& 	3.9854	& 	4.3343 	&	\emph{4.4610}	& 	3.3897	& 	4.1087   \\
MSATr	&	\textbf{3.8944}	& 	3.9550	& 	\textbf{3.9153}	& 	\textbf{4.4023}	& 	\textbf{2.9929}	& 	\textbf{3.8320}   \\

\botrule
\end{tabular}
\end{table}

Generalization performance is significant for deep learning models. Therefore, in this section, the model trained on the LOL dataset mentioned above performs image enhancement testing on five unsupervised datasets. Then, the quality of the generated images is evaluated. As shown in Table \ref{tab3}, except for LIME, MSATr achieved the best results on the other four unsupervised datasets, outperforming all advanced supervised and unsupervised methods on the average of the five datasets. In addition, it can be seen from the experimental results that unsupervised methods are generally better than supervised methods in terms of generalization performance. At the same time, based on data advantages, unsupervised methods have excellent development potential and practical application value. 

\begin{figure}[h]%
\centering
\includegraphics[width=0.9\textwidth]{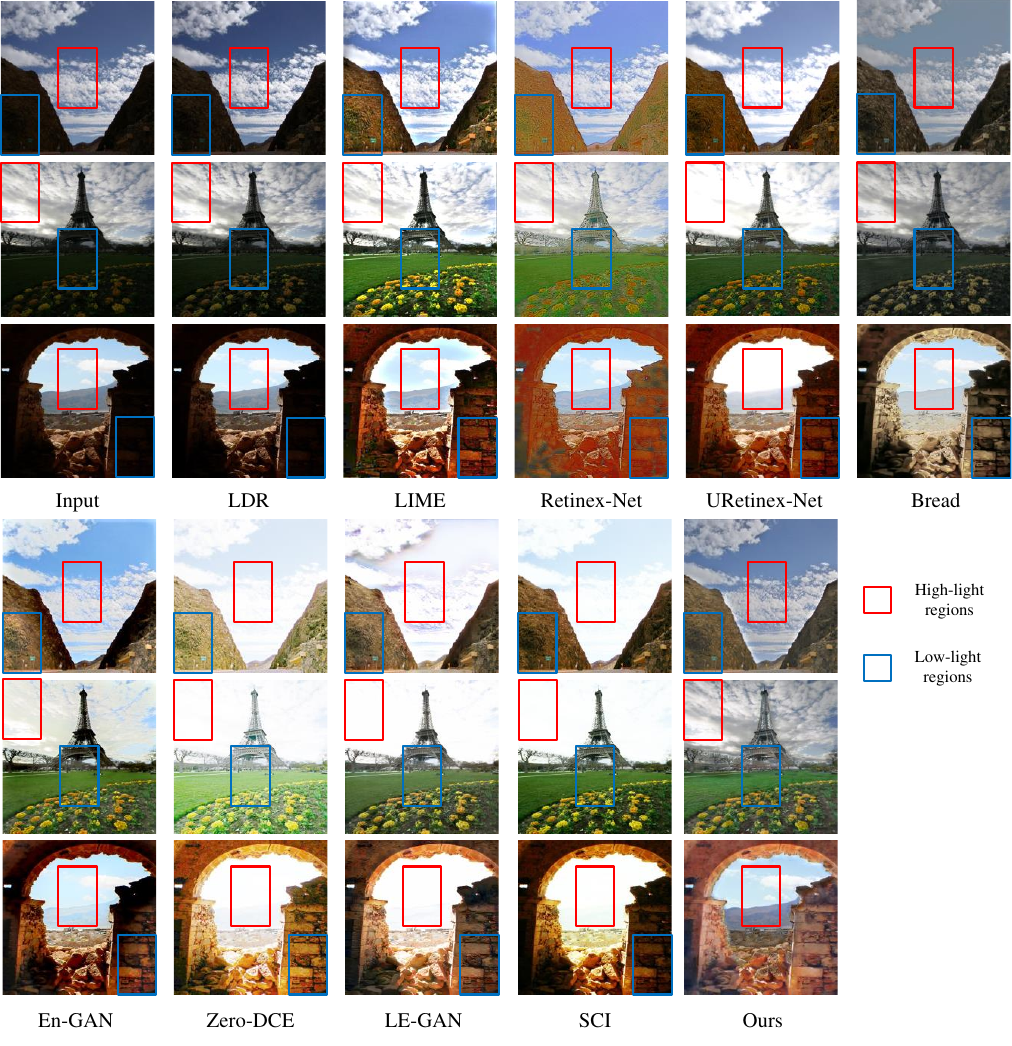}
\caption{Partially enhanced images of each method in the non-reference data set, respectively, from the data sets DICM, MEF, and VV, where the red box represents the highlighted area and the blue box represents the dark area
}\label{fig6}
\end{figure}

Figure \ref{fig6} shows no-reference test images and enhancement effects. The red box represents the bright area of the image, and the blue box represents the dark area of the image. It can be seen that the enhanced appearance of MSATr has the best visual effect. When processing pictures with non-uniform lighting, it can adaptively balance and improve the brightness according to the different brightness between areas, making the overall brightness more natural. It can not only better enhance objects in dark places but also avoid exposure to bright spots. Most other contrast methods have problems of over-enhancement and under-enhancement.

\subsubsection{Analysis of adaptive capabilities of deep learning methods}

\begin{figure}[h]%
\centering
\includegraphics[width=1.0\textwidth]{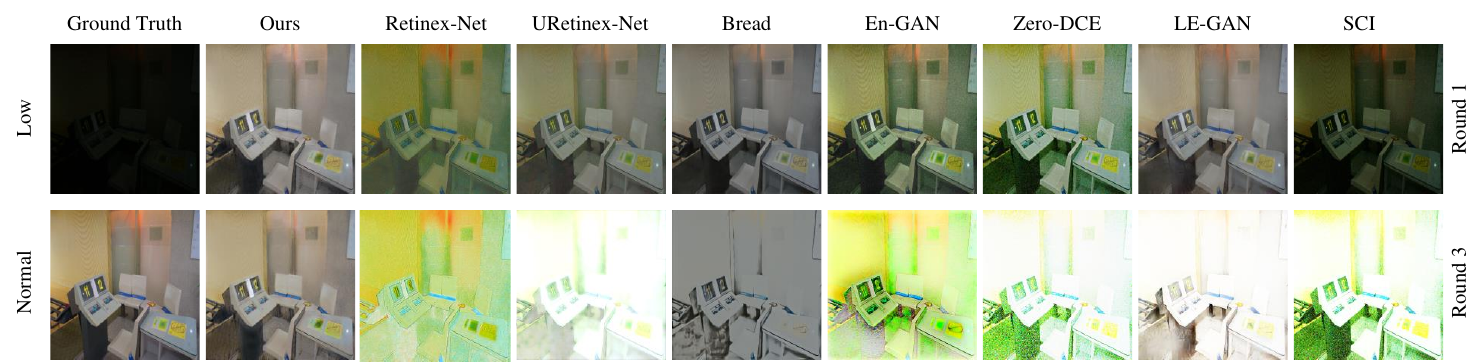}
\caption{Exposure experiments of various deep learning low-light enhancement methods under repeated enhancement.
}\label{fig7}
\end{figure}

This article focuses on adapting the intensity during enhancement and avoiding over-exposure and information loss. We believe the model should always recover the image based on color, luminance, and structure information rather than simply increasing the brightness, especially under non-uniform lighting. Therefore, the experiment in this section analyzes the adaptive perception ability of the deep learning model.  We re-input the enhanced image into the models several times and check for over-exposure. 

Figure \ref{fig7} shows the qualitative results of each low-light image enhancement algorithm for one enhancement and three repeated enhancements. Except for MSATr, all deep learning-based methods produce severe exposure under repeated enhancement, resulting in unacceptable information loss. This means that our MSATr truly realizes the adaptive perception of the brightness of the input image and can perform adaptive enhancement according to the image's characteristics to avoid over- and under-enhancement. 

\subsection{Ablation study}

This section conducts ablation experiments to verify the effectiveness of the local-global structure by retaining a single local attention enhancement network and a global attention enhancement network and retraining them. At the same time, the experiment verified its impact on the network by removing the cyclic training process and luminance consistency loss. The quantitative experimental results are shown in Table \ref{tab4}. On 6 test sets, the NIQE index has declined to a certain extent whether it is a single local or global enhancement network, which shows that the local-global structure can effectively combine its advantages, taking into account the overall and detailed characteristics of the generated image. In addition, after removing the luminance consistency loss, the NIQE index dropped significantly on all data sets, which proves the model's strong brightness perception and excellent enhancement effect.

\begin{figure}[h]%
\centering
\includegraphics[width=0.9\textwidth]{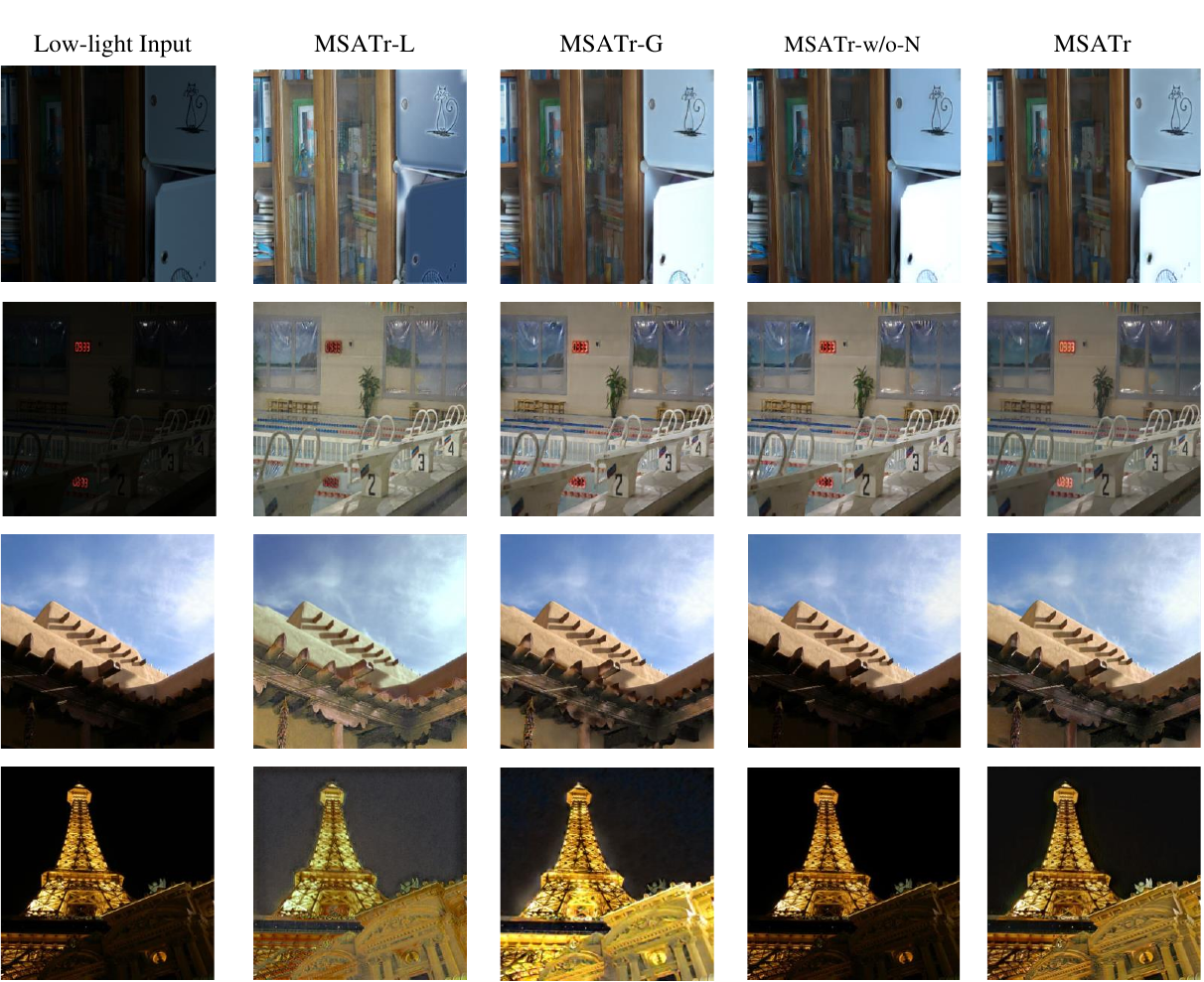}
\caption{Enhanced images of MSATr ablation experiment.
}\label{fig8}
\end{figure}

\begin{table}[h]
\caption{NIQE index of the ablated model on the data set. -L represents the local network, -G represents the global network, and -w/o-N means removing luminance consistency loss. The best results are in bold.
}\label{tab4}%
\begin{tabular}{@{}lccccccc@{}}
\toprule
& \multicolumn{7}{c}{NIQE} \\ \cmidrule{2-8}
Method &	LOL& DICM &	LIME &	MEF &	NPE &	VV &	average\\
\midrule

MSATr- L &	4.0004 & 	4.0517 & 	3.7835 & 	4.0788 & 	4.4489 & 	3.0641 & 	3.9046 \\
MSATr-G &	3.8462 & 	4.2371 & 	\textbf{3.7229} & 	\textbf{3.7438} & 	4.9842 & 	3.2631 & 	3.9662 \\
MSATr-w/o-N &	5.6300 & 	4.4841 & 	4.2410 & 	4.6201 & 	4.8934 & 	3.5996 & 	4.5780 \\
MSATr &	\textbf{3.4813} & 	\textbf{3.8944} & 	3.9550 & 	3.9153 & 	\textbf{4.4023} & 	\textbf{2.9929} & 	\textbf{3.7735} \\

\botrule
\end{tabular}
\end{table}

Figure \ref{fig8} shows the impact of local-global structure and luminance consistency loss of visual quality. It can be seen that the enhanced image of a single local branch network is distorted in brightness and color and cannot balance the brightness of the entire image nicely, but the details are retained relatively clearly. The overall brightness of the generated image with a single global structure is more coordinated, but the details (smaller numbers) are blurry. A complete model trained without luminance consistency loss has difficulty capturing dark areas and performing targeted enhancement when processing non-uniform low-light images, which will affect the adaptive enhancement performance of the entire network for low-light images. The complete MSATr network can synthesize local and global information, ensuring overall brightness coordination and better detail processing. Under luminance consistency loss, over-expose and under-expose will not occur when low-light images are processed.

\section{Conclusion}

This paper introduces an unsupervised low-light image enhancement network MSATr. The multi-scale attention network can better retain pixel-level detailed information through multi-scale window division, self-attention computation and feature interaction. At the same time, a consistency loop training process is used to enhance the model's adaptability for non-uniform low-light images and generalization performance. The brightness of the enhanced image is more balanced, details are more apparent, and the possibility of over-exposure is reduced. The PSNR, SSIM, LPIPS, and NIQE indicators of MSATr in multiple data sets are better than other advanced low-light image enhancement methods, which quantitatively proves the advantages of MSATr over existing methods in low-light image enhancement tasks. The ablation experiment verified the working characteristics of the local-global transformer structure and the effectiveness of the loop training strategy. 

\backmatter
\bmhead{Acknowledgments}

The authors would like to thank their colleagues from the machine learning group for discussions on this paper. This work was supported by Science \& Technology Project of State Grid Corporation of China (No.5400-202355230A-1-1-ZN)

\bmhead{Author Contributions}

Xiao Fang: Methodology, Software, Writing - Original Draft, Writing - Review \& Editing.\\ Xin Gao: Conceptualization, Methodology, Supervision, Writing - Original Draft, Writing - Review \& Editing.\\ Baofeng Li: Software, Validation, Funding acquisition.\\ Feng Zhai: Conceptualization, Resources, Funding acquisition.\\ Yu Qin: Software, Validation, Funding acquisition.\\ Zhihang Meng: Writing - Review \& Editing.\\ Jiansheng Lu: Conceptualization, Resources, Funding acquisition.\\ Chun Xiao: Conceptualization, Validation, Funding acquisition.

\bmhead{Availability of data and materials}

The datasets supporting the results of this article are LOL, DICM, LIME, NPE, MEF and VV public datasets, and the authors confirm that the datasets are indicated in reference list.

\section*{Declarations}


\begin{itemize}
\item Competing interests: The authors have no competing interests to declare that are relevant to the content of this article.
\item Ethics approval: Not applicable. 
\item Consent to participate: Not applicable.
\item Consent for publication: Not applicable.
\end{itemize}

\bibliography{export}

\end{document}